# Can A Cognitive Architecture Fundamentally Enhance LLMs? Or Vice Versa?

Ron Sun

Department of Cognitive Science
Rensselaer Polytechnic Institute
Troy, NY 12180, USA
dr.ron.sun@gmail.com

## Abstract

The paper discusses what is needed to address the limitations of current LLM-centered AI systems. The paper argues that incorporating insights from human cognition and psychology, as embodied by a computational cognitive architecture, can help develop systems that are more capable, more reliable, and more human-like. It emphasizes the importance of the dual-process architecture and the hybrid neuro-symbolic approach in addressing the limitations of current LLMs. In the opposite direction, the paper also highlights the need for an overhaul of computational cognitive architectures to better reflect advances in AI and computing technology. Overall, the paper advocates for a multidisciplinary, mutually beneficial approach towards developing better models both for AI and for understanding the human mind.

## Introduction

AI systems, especially LLMs, trained on a massive amount of data and with a huge number of parameters, have achieved spectacular successes in recent years, in terms of, for example, natural language generation or human-level test performance. These systems have found their way into large-scale, real-world applications, in domains ranging from creative writing to autonomous driving, dealing with languages, images, voices, and so on.

However, at the same time, unlike humans, current LLM-centered systems (as well as other deep learning models) suffer from some fundamental shortcomings, including limited abstract reasoning or planning capabilities, limited memory, lack of autonomy, lack of human-like generalization, limited reliability and trustworthiness, and so on. For instance, such systems are often not reliable --- producing content that is meaningless or simply false. They may be inconsistent with human preferences, deviating from human values or ethics. When faced with complex problems, LLMs may have difficulties in achieving human-level performance, by way of gathering relevant information, reflecting on successes and failures, decomposing and recomposing problems, and using various cognitive strategies that humans seem to possess.

These shortcomings often stem from not taking human psychology into serious enough consideration in LLMs. We thus should explore how the understanding of the human mind (e.g., from cognitive science and psychology) can help in improving them. This can lead to future systems that are more trustworthy and more aligned with human understanding, human values, and so on (beyond current methods such as reinforcement learning from human feedback). Future systems achieving human-level performance should, almost by necessity, draw inspirations from human psychology and models of the human mind (from computational psychology; Sun, 2023). In the ongoing discourse regarding the limits of LLMs, the debate is often between those who think that more data alone can provide the solution and those who favor incorporating additional methods. From the afore-mentioned perspective, the latter position seems more plausible, but nevertheless it needs to be strictly guided by empirically grounded understanding of the human mind.

That is to say, future systems should be more rigorously structured in a cognitively (psychologically) motivated and justified way, based on the structure and content of human mental processes. *Computational cognitive architectures* that capture (to some significant extent) human mental processes do exist --- They are empirically (psychologically) grounded, comprehensive computational frameworks for understanding and potentially capturing the human mind (e.g., Sun, 2016). In such architectures, *dual processes* should be considered, as it has become increasingly evident that they are a key feature of the human mind (Kahneman, 2011; Reber, 1989; Sun, 1994, 2002, 2015). Dual-process computational cognitive architectures tap into psychological details of implicit (roughly, unconscious, intuitive) versus explicit (roughly, conscious, deliberative) psychological processes (i.e., System 1 versus 2).

I argued a long time ago (e.g., Sun, 1994) as well as very recently that hybrid neuro-symbolic systems seem necessary to address dual processes, which can also remedy many existing shortcomings of LLMs. That is, to build systems that achieve human-level flexibility, reliability, and robustness, one may need both symbolic and subsymbolic (neural) methods, with each side contributing different capabilities. For example, the subsymbolic side may address lower-level sensory-motor processes and unconscious (implicit) processes. The symbolic side can perform high-level "conscious" reasoning, explicit planning, and deliberative reflection. These two sides may be distinct, but work together synergistically (Sun et al., 2005). Such neuro-symbolic models for tackling dual processes have been worked on since the 1990s (e.g., Sun, 1994). More theoretical arguments can also be found in Sun (2002).

Furthermore, another aspect at least as important as dual processes is essential or intrinsic human needs or motives, which are key elements of the human mind (Murray, 1938). They are the basis of directing and regulating behavior. This aspect is well developed in some cognitive architectures. They are important because they are central to alignment with human intentions, values, and ethics (Bretz & Sun, 2018), as well as for dealing with social interaction (Sun, 2006). Of course, many other aspects of the human mind are also important to LLMs, as will be explicated later.

On the other hand, looking at the issues from the opposite direction, we see that the field of computational cognitive modeling, including computational cognitive architectures, has been in development since at least the 1970s. It is concerned with developing models of the human mind in a psychologically realistic way. Although these models are meant to capture human performance in a range of activities, their actual capabilities have not been keeping up with the advances in new technology. For instance, some cognitive architectures were initially conceived in the 1970s and relied on technology available at that time (outdated by now). Despite the fact that some new ideas and implementations have been incorporated, their overall frameworks, as well as some of their essential details, nevertheless seem out of date fundamentally. Therefore, it may be argued that computational cognitive modeling and cognitive architectures need an overhaul to incorporate, more readily and more seamlessly, new advances in AI and in computing technology. LLMs may serve as the key to this endeavor due to their comprehensive capabilities (as will be elaborated later). Note that here we are concerned with psychologically realistic computational cognitive architectures that are meant to be rigorous models of the human mind and have been validated empirically.

In the remainder of this article, I will investigate what LLMs are fundamentally capable of and what else may be needed. I then argue that computational cognitive architectures may help to enhance LLMs, and a detailed example is presented. Several key aspects are discussed and weaved together.

## What Can LLMs Capture?

There are many different LLMs, so it may be hard to generalize. Risking being a little simplistic or over-generalizing, one may consider the notions of intuition and instinct (Cosmides &Tooby, 1994; Dreyfus & Dreyfus, 1986; Sun & Wilson, 2014) in characterizing LLMs (besides implicit processes in general; more on these later). Here, intuition refers to understanding without the involvement of explicit reason (but regardless of whether it is immediate or slow in emergence; Sun, 2015), while instinct refers to relatively fixed patterns of responses to environmental stimuli without involving explicit reason (McFarland, 1989; Sun & Wilson, 2014). One may claim that LLMs correspond roughly to human intuition and instinct (setting aside perception, motor control, language, and other peripheral implicit processes for the time being).

Why should LLMs correspond to intuition and instinct? What characteristics of LLMs make them correspond to intuition and instinct? Let us look into intuition first in some detail.

Human intuition results from implicit (unconscious) processes (Evans & Frankish, 2009; Kahneman, 2011; Reber, 1989). Intuition has often been defined as ''the immediate apprehension of an object by the mind without the intervention of any [explicit] reasoning process'' (Oxford English Dictionary), or ''immediate apprehension or cognition'' (Merriam-Webster Dictionary). One may view intuition as a form of reasoning: Reasoning encompasses explicit processes on the one hand, and implicit processes (intuition) on the other (Sun 1994), both guided by one's motives, needs, and goals. In fact, intuition, as well as insight resulting from intuition, is arguably indispensable elements in the overall process of human thinking and reasoning --- They supplement and guide explicit reasoning (Helie & Sun, 2010). Prior empirical work demonstrating the above can be found in Evans and Frankish (2009), Reber (1989), and so on, while theoretical work (with computational modeling) includes Sun (1994, 2015), Helie & Sun (2010), and so on.

Admittedly, intuition might be, though not necessarily always, beyond what language can express; at least, the process leading up to intuition is often beyond language (Dreyfus & Dreyfus, 1986; Sun, 1994). However, intuition is, at a minimum, partially expressible through language (even though it might not necessarily be completely expressed by language). Given an extremely large amount of linguistic data, it is likely that human intuition about a vast array of subjects may be contained in (or embodied by) it. With an extremely large amount of linguistic training data, it may be argued that LLMs resulting from such training can gain a large range of intuition, somewhat comparable to human intuition. That is, although an individual piece of intuition might not be captured by an individual sentence, a vast number of sentences from a vast number of different sources can conceivably capture a fair amount of intuition collectively. Furthermore, human intuition results from and relies on diverse (likely implicit) common-sense knowledge about how the world works (including, e.g., naive physics, folk psychology, folk sociology, and so on), which is what an extremely large amount of linguistic training data from diverse sources can provide. On top of that, the numerical training process with gradual numerical weight adjustments in training neural networks (including in training Transformers, the basis of LLMs) leads to generalization (including both interpolation and extrapolation), which in turn leads to better intuition of an even broader scope. Thus, conceivably, LLMs can embody human intuition as gleaned and generalized from a large volume of data.

Looking at this issue in another way, in humans, intuition is often obtained through experience, especially a large amount of repeated experience. On the other hand, a large amount of repeated experience is exactly what the training of LLMs with a large amount of linguistic data provides. Through such experience, human-like intuition develops within LLMs. Some may claim that there is a fundamental difference between intuition and language expressing intuition, or between intuition gained from the "real" world and whatever gained from linguistic texts. This claim seems to build a wall separating language and the "real" world unnecessarily and unjustifiably. To any individual, the world is made up of sensory-motor, linguistic, and other experiences, which are closely entangled and often mirroring each other (Pavlik, 2023). Word meanings may emerge from use patterns (Durt et al., 2023), so a proximate function such as next-token prediction may give rise to an ultimate function such as intuition (Mollo & Millière, 2023). One may gain intuition from any or all of these modalities (see "grounding" later).

Furthermore, human intuition may result from grasping underlying statistical patterns and structures of the world in various respects in an implicit way, due to repeated experiences (Hasher & Zacks, 1979; Reber, 1989). Likewise, repeated training of LLMs with a large amount of data facilitates the (implicit) capturing of statistical patterns and structures of the data within LLMs (Durt et al., 2023; Mollo & Millière, 2023; Zhang et al., 2023). For example, Dasgupta et al. (2022) and Saparov & He (2022) showed that LLMs demonstrated content effects in reasoning similar to humans. Trott et al. (2023) showed that LLMs could capture human intuition exhibited in false-belief experiments. And so on.

One characteristic of human intuition is that one often makes quick "inferences" without (consciously) considering all possible alternatives. In other words, one often automatically (unconsciously) filters out what is (perceived to be) unlikely or less likely and thus focuses on the most probable inference only (Dreyfus & Dreyfus, 1986). Moreover, when communicating an intuition in language, one may express it one word/phrase at a time on the fly (in a way that resembles "thinking aloud" protocols in psychological experiments), that is, vocalizing thoughts as they unfold, apparently without necessarily a plan for sentence structures or wording in advance. The two aspects above are highly similar to the way in which LLMs generate sentences and paragraphs (although LLMs can be instructed to do otherwise to some limited extent; e.g., Wei et al, 2022). This is probably because, in humans as well as in LLMs, the size of the set of all possibly inferences and/or the size of the set of all possible verbal descriptions of an intuition are both too large for humans and for LLMs to enumerate and examine deliberatively one by one in real time. Thus selectivity becomes necessary for humans and LLMs alike when dealing with real-time, on-the-fly generation of inferences or verbal descriptions.

But, assuming that LLMs can capture intuition, can LLMs capture more than intuition? For one thing, human language involves structured and/or symbolic processes, judging from the output of the human language faculty (as argued by, e.g., Fodor & Pylyshyn, 1988; Pinker & Prince, 1988), which seems to suggest symbolic mental capabilities. LLMs do likewise have some symbolic capabilities, including structures and compositionality to some extent (e.g., Pavlick, 2023), especially when it comes to dealing with language production and comprehension. However, it is worth pointing out that LLMs (at least the currently existing ones) have only limited such capabilities. At the same time, human implicit mental processes have been shown to have likewise limited such capabilities (Macchi et al., 2016; Reber, 1989), when explicit processes are not involved. When explicit mental processes are involved, they can lead to a full range of explicit symbolic capabilities; for example, full explicit symbolic reasoning can be exhibited by humans. Therefore, it seems unnecessary to posit in LLMs anything more than implicit processes to explain limited symbolic capabilities of LLMs.

This view is consistent with some views expressed in the literature. Bubeck et al. (2023) assessed performance of GPT-4 on a variety of tasks and concluded that it could capture "fast" (intuitive, implicit) processes, but not "slow" (reflective, explicit) processes. Mugan (2023) likewise asserted that LLMs captured implicit, unconscious processes. Pavlik (2023) argued that processes within LLMs, although implicit, could perform some apparently symbolic operations and inferences.

Now, beside intuition that was just discussed, what about another major type of implicit processes --- instinct? Instinct is the ''tendency of an organism to make a complex and specific response to environmental stimuli without involving [explicit] reason'' (Merriam-Webster Dictionary). One may also link it to the Heideggerian notion of "comportment" (Dreyfus & Dreyfus, 1987). Empirical work has been done to elucidate this concept: For instance, ethologists have studied animal instincts extensively (McFarland, 1989). Some work has also been done to elucidate this notion through computational models: Sun & Wilson (2014) discussed how human motivation and consequent implicit action selection (along with other processes) together capture instinct. It is mostly implicit and centers on the interaction of internally felt needs (motives) and external environmental factors in (implicitly) determining actions, thereby capturing much of characteristic behavioral patterns (e.g., personality; Sun & Wilson 2014). Beyond hardwired (innate) instincts, there are also acquired (learned) instincts, mostly through repeated experiences. To both kinds of instincts, what was argued earlier regarding intuition can be largely applied. In other words, instincts can develop within LLMs through repeated training with a large amount of data (more on this later). Arguments are somewhat similar and thus will not be repeated.

Note that different types of implicit mental processes, such as intuition versus instinct, may be captured by different and separate LLMs, as they have been structured in existing psychological models of intuition and instinct by parallel modules or parallel subsystems (e.g., Sun & Wilson, 2014).

## What Else is Needed Beyond LLMs?

In order to achieve human-level intelligence or to understand and capture the full capacity of the human mind, what else do we need beyond the currently existing LLMs? To answer this question, we can take cues from cognitive science, psychology, neuroscience, and so on: Theoretical and computational models and empirical findings from these disciplines can help. In particular, as mentioned before, computational cognitive architectures developed therein can be illuminating on this question.

From a vast amount of research in cognitive science, psychology, and neuroscience, we know that humans are capable of explicit (conscious) cognition involving, among other things, symbol manipulation and explicit rule following (Fodor & Pylyshyn, 1988; Pinker & Prince, 1988; Sun, 1994), beyond mere implicit processes (which may be captured by neural networks, e.g., by LLMs, as argued so far). Therefore, explicit processes that involve symbol manipulation need to be included in a more complete model of intelligence, especially in a psychologically realistic model of human intelligence.

In this regard, dual-process theories, formalizing and codifying what has been stated above, are worth noting. They have been gaining attention in recent decades, in psychology, philosophy, and many other fields. The distinction between implicit processes (also termed System 1 or simply "intuition") and explicit processes (also termed System 2 or "reason") has been argued by many since the 1980s (e.g., Reber, 1989; Sun, 1994, 2002) and popularized later by Kahneman (2011). In general, explicit processes are relatively easily accessible to consciousness, while implicit processes are less so. Explicit processing may be described as symbolic and rule-based to some significant extent, while implicit processing is more ''associative'' and "holistic" (Sun, 1994, 2002). That is, explicit processing may involve the manipulation of symbols, while, in contrast, implicit processing involves more instantiated knowledge that is holistically and/or statistically associated. Empirical evidence and theoretical analysis in support of these points can be found in the literature (e.g., in the work cited above). Work on computational cognitive architectures has also demonstrated computationally the importance of dual processes (e.g., Helie & Sun, 2010; Sun et al., 2005).

Furthermore, beyond implicit versus explicit processes, a number of other aspects that have usually been neglected in LLMs are also crucial. For instance, motivation is of fundamental importance to humans and human-like behavior, as has been argued before (e.g., Maslow, 1943; Reiss, 2004; Ryan & Deci, 2000; Sun et al., 2022).

Specifically, if one does not have goals (or has only randomly set goals that are randomly changing), one will inevitably act in a disorganized, haphazard way; or one may rely only on fixed reflexes that are rigid and inflexible (namely, much less intelligent). Similarly, for AI systems, without motivational processes, they would be aimlessly; or they would have to rely on prior knowledge coded into them (just like reflexes in humans), in order to accomplish something in only relatively simple circumstances. Alternatively, they may rely on external "feedback" to learn to perform right actions (using, e.g., reinforcement learning). But the requirement of external feedback begs the question of how it should be obtained in the natural world.

Largely absent in current AI systems are essential (or intrinsic) motivation, as well as its correlates: emotion, personality, and other human characteristics (Sun & Wilson, 2014). They are important, not just for attaining goal-orientedness and complex behavior (as discussed so far), but also for addressing the meta-level issues of alignment with human intentions, values, and ethics (Bretz & Sun, 2018), as well as for competency in dealing with social interaction (Sun, 2006). Motivation should be an important part of any human-level system, in order to account for deeper structures in the control of behavior. In this regard, Maslow (1943), Murray (1938), Reiss (2004), and Sun et al. (2022) showed the complexity of human motives and needs, which may serve as roadmaps for enhancing LLMs. In addition, correlated with motivation, personality, emotion, culture, and so on also need to be taken into consideration, both to capture human-like behavior and to understand the human mind. Regarding their respective roles in cognitive architectures, see Sun and Wilson (2014), Sun et al. (2016), and so on.

There are several other general ideas of what current LLMs lack but need (e.g., metacognition, memory, online learning, etc.; more on them later). In addition, beyond general ideas, one may delve into details of existing LLMs for more fine-grained understanding of what is needed. At a more detailed level, one may ascertain the capabilities and shortcomings of existing LLMs (e.g., Bubeck et al, 2023; Chang & Bergen, 2023; Pavlick, 2023); this may be achieved, in part, through applying a variety of psychological and other tests to LLMs (e.g., Binz & Schulz, 2023; Trott et al., 2023). With a more precise understanding, more fine-grained ways of enhancing LLMs may be devised.

## Can a Cognitive Architecture Be the Right Framework?

Can a computational cognitive architecture, as mentioned earlier, help in a general way and serve as an overarching framework for the sake of enhancing and strengthening LLMs? The answer so far seems to be yes. For instance, computational cognitive architectures can make it possible for an LLM-based agent to better remember and retrieve information, choose actions in dynamic (physical and social) environments, plan future courses, reason about and solve problems, reflect metacognitively, and interact effectively with other agents (human or otherwise), as directed by one's intrinsic motives and needs.

This is not entirely a new idea: Some have already thought of cognitive architectures in a similar way, albeit very recently (and sometimes not very "cognitive"). Xie et al. (2023) drew on existing cognitive architectures for enhancing LLMs, thereby incorporating attention, memory, reasoning, learning, and decision-making mechanisms. Park et al. (2023) described an architecture that extends an LLM to store an agent's experiences using natural language, distill those (linguistic) memories into higher-level reflections, and retrieve them to plan behavior. Romero et al. (2023) discussed several specific possibilities in this regard. Sumers et al. (2023) surveyed more broadly.

With a cognitive architecture, an agent takes its current environmental information, past experiences, and internal needs and goals into account when generating behavior. Under the hood, the cognitive architecture may combine a LLM with various mechanisms for organizing and retrieving relevant information as well as for self-regulation and focusing on important objectives, to help the LLM to generate useful outputs and to maintain behavioral consistency over time. Both implicit and explicit processes are involved (according to some cognitive architectures). In particular, the interaction between implicit and explicit processes within a cognitive architecture (which has been extensively studied; e.g., Sun, 2016) can be leveraged for structuring the interaction between LLMs and symbolic processes.

In the other direction, LLMs can serve as an important tool and an underlying technology that elevate existing cognitive architectures beyond laboratory toys towards both theoretical and practical tools for the real world. LLMs better addresses real-world complexity, given the vast amount of data used in training LLMs, as demonstrated by many use cases seen thus far. LLMs can also lead somehow to better psychological realism of cognitive architectures, in the sense that they may generate psychologically plausible behavior in complex environments (beyond typically small laboratory experiments used in validating cognitive models). Psychological realism of LLMs is being actively explored by researchers (Dasgupta, et al., 2022; Trott, et al., 2023; etc.).

## An Example of a Dual-process, Dual-representation, and Motivation-focused Cognitive Architecture

I will use the Clarion cognitive architecture as an example. Clarion is meant to be a comprehensive, mechanistic, process-based, psychological theory, with detailed computational specifications and implementations (Sun, 2002, 2016).

Clarion consists of four major subsystems: the action-centered subsystem (ACS) for dealing with actions involving procedural (i.e., action) knowledge, the non-action-centered subsystem (NACS) for reasoning and memory involving declarative (i.e., factual) knowledge, the motivational subsystem (MS) for dealing with motivation, and the metacognitive subsystem (MCS) for regulating other subsystems (Sun, 2016).

Each of these subsystems consists of two "levels". The top level carries out explicit (roughly, conscious, deliberative) processes; the bottom level carries out implicit (roughly, unconscious, intuitive) processes (Reber, 1989; Sun, 1994, 2002). Computationally, one is symbolic and the other neural. The two types of representations are inter-connected: The symbolic representations at the top level, which are in the forms of "chunks" (each representing a concept, defined by a set of dimension-value pairs) and rules connecting chunks, are linked to corresponding neural representations at the bottom level. The two processes thus interact to generate combined outcomes. See Figure 1.

The flow of information among these subsystems is roughly as follows: First, within the MS, situational inputs trigger internal (mostly intrinsic) motives, termed *drives*, based on both internal propensities and situational inputs. Then, a goal (an intention for action) within the MS is selected (by the MCS) on the basis of activated motives/drives (for the sake of satisfying these). Action/meta-action selection occurs, within the ACS/MCS, on the basis of the goal selected and the situational inputs (to maximize the satisfaction of the motives/drives).

Note that Clarion has been well validated against psychological data, findings, and theories (Sun, 2002, 2016). It constitutes a comprehensive theory of the mind. It is suitable as a framework for enhancing LLMs (as partially explored already, e.g., by Romero et al., 2023).

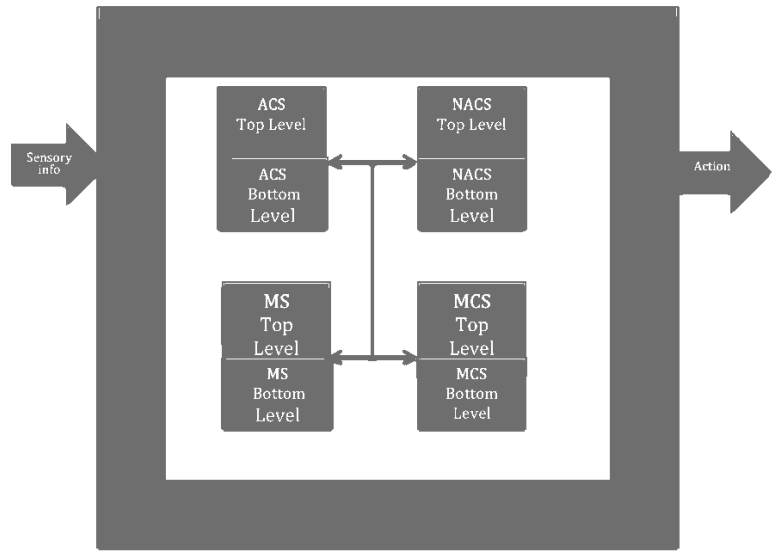

Figure 1. The original Clarion cognitive architecture with its four major subsystems (from Sun, 2016).

## How Can LLMs be Incorporated into Clarion?

For incorporating LLMs into Clarion, there are multiple possibilities, ranging from LLMs being at the periphery of the cognitive architecture to LLMs being at its very core.  Here are five possibilities for LLMs in Clarion (see Romero et al., 2023 for some of these and other alternatives):

- For dealing with perception: Multimodal LLMs handle (implicit early stages of) perception processes for Clarion. Given an input (e.g., a visual scene), LLMs generate linguistic and/or structured descriptions (in a form compatible with the internal representations of Clarion), for processing and then responding by Clarion.

- For dealing with language-based communication: In this case, natural language inputs are directed to LLMs and LLMs (implicitly) process them to provide inputs to Clarion (in a form compatible with the Clarion internal representations). Clarion can then respond to them in its internal forms, and LLMs generate natural language texts on that basis, for the sake of verbal communication (although full natural language understanding would likely require more than that).

- For dealing with motor processes:  LLMs (implicitly) generate motor actions, based on directives from Clarion (in an internal form) that provide general specifications for the motor actions.

- For capturing memory: In this case, some modules of Clarion may be based on LLMs, which serve as (implicit) memory of various forms in Clarion (e.g., implicit semantic or procedural memory).

- For capturing all forms of implicit processes: LLMs can capture all implicit processes, ranging from intuition to instinct (as argued before), aside from (implicit processes of) perception, motor control, and natural language communication (as mentioned above). In this case, LLMs play all these roles.

The last approach above evidently confers the most extensive role on LLMs within a cognitive architecture, as it encompasses all other possibilities essentially; for example, perception, motor, and language processes are all largely implicit. Other modules or components within a cognitive architecture therefore must work closely with LLMs.

Among these possibilities, this last approach is preferrable for its generality. This approach is also well justified theoretically (see the discussion earlier on dual processes). Moreover, this approach is readily workable in any cognitive architectures that incorporate dual processes. In other words, LLMs are a natural fit for adding to and enhancing such cognitive architectures and, conversely, such cognitive architectures are well positioned to strengthen LLMs as well.

As will be detailed, it can be argued that the overall architecture of Clarion and many of its technical details are applicable for the sake of *structurally* and *fundamentally* enhancing LLMs. Such enhancement is not just adding more parameters to LLMs, not just changing their pre-training methods, and not just fine-tuning them (after pre-training), but much more fundamental as it incorporates well-established elements of the human psychology that are otherwise missing in current LLMs.

Below I investigate this approach in some detail. See Figure 2 for a sketch of the overall architecture, as will be explicated below. Note that, due to the length limit, not all aspects can be covered, but only a few examples of key aspects.

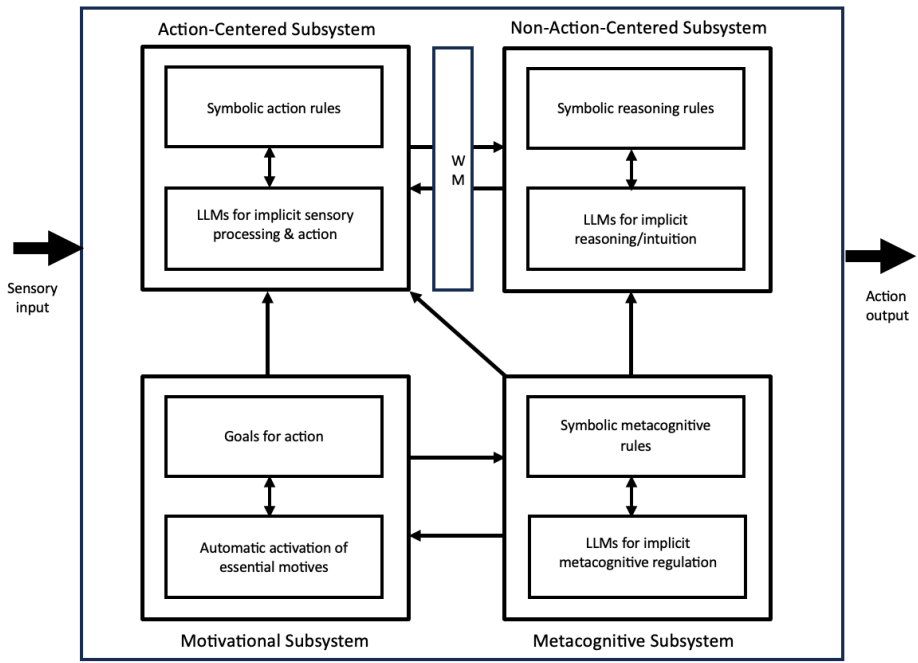

**Figure 2. The Clarion architecture with LLMs incorporated.**

## Human-like Dual Processes: Explicit, Symbolic Processes Interacting with LLMs

As argued earlier, dual processes must be taken into serious consideration, in order to enhance and strengthen LLMs: We need to include what correspond to implicit processes (such as those computationally captured by neural networks, including LLMs) and what correspond to explicit processes (such as those computationally captured by symbolic and/or rule-based systems). Concerning LLMs specifically, Mugan (2023) pointed out the need for "tools for deliberate thinking". Bubeck et al. (2023) also noted the missing "slow-thinking" (i.e., explicit) component in LLMs, believing that the "fast-thinking" (i.e., implicit) component could serve as its subroutine.

However, each type of process may not be just tools; rather, each is a fundamental part of the human mind (Sun, 1994, 2002). Further, explicit processes may guide (affect in some way) implicit processes, but at the same time, they may also be guided (affected) by implicit processes (Sun, 2002). Thus, complex and dynamic interaction occurs between these two types (in notable contrast to some existing theories, such as the "default-intervention" view of Evans & Frankish, 2009). It should be noted that somewhat similar ideas, although much courser-grained (i.e., not yet addressing finer psychological details), have been proposed by Booch et al. (2021), Lin et al. (2023), and others, with one or more modules consisting of LLMs.

In Clarion, detailed processes of implicit-explicit interaction have been elucidated mechanistically (i.e., computationally; in the ACS and the NACS; Sun, 2016), and validated psychologically (to a significant extent; through modeling and simulation of empirical psychological data). These interaction processes can be mapped, when LLMs are incorporated into Clarion, to those between LLMs (capturing implicit processes) and explicit, symbolic processes. That is, the psychologically validated dual processes within Clarion can be leveraged for structuring the interaction between LLMs and symbolic processes.

Within Clarion, one essential form of interaction is what has been termed *top-down activation* (initiation of implicit processes by explicit processes), while the other is *bottom-up activation* (activation of symbolic structures by implicit processes). Thus, for instance, implicit processes may be guided by explicit processes (e.g., by explicit instructions), and results of implicit processes may be confirmed, refined, rectified, or explained by explicit processes that are in turn grounded in implicit processes. These two forms are crucial to Clarion.

In relation to learning, one form of implicit-explicit interaction in Clarion is *top-down learning* (taking explicit knowledge and assimilating them into implicit forms), and another is *bottom-up learning* (learning implicit knowledge first and then learning explicit knowledge on that basis). See Sun (2002, 2016) for technical details of these forms, which enable incremental, continuous, real-time learning specific to situations encountered (more on learning later).

With LLMs capturing implicit processes, top-down and bottom-up activations originally specified in Clarion become (1) prompts to LLMs, and (2) outputs from LLMs in a form that can be linked up with symbolic processes at the top level of the architecture. Note that outputs from LLMs can be in linguistic or Clarion internal representations (depending on prompts to LLMs and other factors). As noted earlier, internal representations in Clarion are in the form of chunks (each defined by a set of dimension-value pairs, serving as a condition or a conclusion in explicit reasoning), or rules (with their conditions and conclusions specified by chunks; acquired through learning, e.g., bottom-up learning). (For now, I ignore utilizing embeddings from LLMs, which would require a much longer treatment.) On the other hand, inputs to LLMs are prompts (in a linguistic or an internal form), along with possibly other information. Thus prompt engineering and other techniques are needed to get desired outcomes from LLMs. It is worth noting that LLMs can translate between linguistic (natural language) representations and internal representations used in Clarion, which facilitates the processing of external natural language inputs (e.g., queries, instructions, or situational descriptions).

In Clarion with LLMs incorporated, there are several roles for explicit, symbolic processes at the top level: (1) providing prompts to the LLMs at the bottom level to initiate their processing; (2) verifying, rectifying, refining, or explaining the results returned from the LLMs (which correspond to intuition or instinct) at the top level, using explicit knowledge at that level; (3) supplementing or even supplanting the results from the LLMs, using explicit knowledge at the top level, when situations call for it. Within the context of Clarion, prompting, in a way, amounts to an inner dialogue within oneself (Vygotsky, 1962). In this case, the dialogue mainly consists of repeated queries to one's inner self (i.e., intuition and so on). Through this dialogue, one's thoughts form, develop, and mature.

In a way, the Clarion approach of dual representations achieve similar outcomes as Instruction Tuning (or Agent Tuning; cf. Zeng et al., 2023) would eventually. But this approach is less time consuming (e.g., no repeated tuning), is more flexible, and can instantaneously change to adapt to new situations.

Below I will examine dual processes in two cognitive functions.

## Reasoning

Reasoning is a characteristically human capability. Dasgupta et al. (2022), Saparov & He (2022), Trott et al. (2023), Zhang et al. (2023), and others assessed reasoning by LLMs and found mixed results. Bubeck et al. (2023) pointed out that LLMs suffered from "lack of planning, working memory, ability to backtrack". In other words, although LLMs may perform some (intuitive) reasoning, they are not yet capable of full human-level reasoning.

One may argue that various ideas in terms of "prompt engineering" together show the importance of explicit, symbolic processes in guiding LLMs for reasoning (through one prompt or through iterative prompts). For example, Chain of Thought, Tree of Thought, and Algorithm of Thought prompts (e.g., Wei et al., 2022) provide one-prompt stepwise guidance from explicit to implicit processes. In additional, common search control strategies, such as depth-first, breadth-first, and A* search, can be implemented from explicit processes in Clarion (e.g., through multiple, iterative prompts). Furthermore, prompts to LLMs can involve complex, sophisticated reasoning templates, which again are guidance from explicit to implicit processes: Xie et al. (2023) showed that reasoning in LLMs could be enhanced through templates (given by explicit processes) for performing analogical reasoning, problem decomposition, planning thinking, integrative thinking, and so on. Xi et al. (2023) surveyed additional methods. In the original Clarion, there are a number of other reasoning methods (templates), such as forwarding chaining, backward chaining, constraint satisfaction, and so on (see Sun, 2016), which can be invoked by explicit processes (especially from the MCS; Sun, 2016). So, explicit processes can guide LLMs, in addition to the fact that they can perform explicit, symbolic reasoning with precision, certainty, and other useful characteristics in conjunction with or instead of (supplanting) LLMs. Together, the implicit-explicit interaction in Clarion leads to much enhanced reasoning abilities, combining intuition and explicit reasoning.

Reasoning also relies on various (implicit or explicit) mental models. These models may be formed through (semantic, episodic, and procedural) memory and learning, as will be discussed later.

### Action

Transformers or LLMs can capture human-like instincts for actions as discussed earlier (Sun & Wilson, 2004). Chen et al. (2021) proposed that one could use Transformers to capture action sequences, just as they learned natural language sentences. The model, when provided partial action sequences occurred thus far, predicted the next action to be taken. Lin et al. (2023) and others proposed methods that turn outputs of regular LLMs into actions (e.g., through action templates).

Thus, within Clarion, essentially the same kind of model (LLMs based on Transformers) can be used to capture a variety of implicit processes, including both intuition and instinct (as discussed before), in addition to language, perception, and motor functions. Note that implicit processes for reasoning and action are captured by separate but interacting LLMs (cf. Yao et al., 2023), corresponding exactly to the distinction of and the relation between the ACS and the NACS in Clarion (in their bottom levels).

However, in Clarion, on top of implicit action selection processes, explicit, symbolic processes are also present. Besides providing prompts to LLMs (e.g., as guidance), they also provide precision, certainty, and other useful characteristics, in conjunction with or instead of (supplanting) LLMs.

In addition, planning of multi-step action sequences (an essential human capability) can also be carried out in Clarion through implicit-explicit interaction. For instance, Sun and Sessions (2000) showed that explicit, symbolic plans can emerge from neural reinforcement learning. Recent work involving planning through LLMs (through prompts to initiate planning in LLMs) includes Lin et al. (2023), Park et al. (2023), Yao et al. (2023), and so on. Details of planning are omitted, due to space constraints.

### Human-like Memory: Mnemonic Aid to LLMs

In Clarion, there are a variety of memory: semantic memory, procedural memory, episodic memory, working memory, and so on, often in both implicit and explicit forms. These types of memory are crucially important for human cognition, as has been demonstrated time and again in empirical psychological research (e.g., Baddeley, 1986; Schacter, 1987; Tulving, 1983). They are thus also important to Clarion in capturing human cognitive processes (Sun, 2012).

LLMs, given their capabilities, serve well as implicit semantic memory and implicit procedural memory within Clarion (at the bottom levels of the NACS and the ACS, respectively), while the top levels of the NACS and the ACS serve as explicit semantic and procedural memory (Schacter, 1987; Sun, 2012). However, episodic memory (Tulving, 1983) and working memory (Baddeley, 1986) are currently lacking in most LLMs. Clarion provides useful insights as to how to incorporate these needed memory systems, so as to lead to enhanced performance overall (Sumers et al., 2023; Xi et al., 2023).

In Clarion, explicit episodic memory consists of an agent's experiences at each time step, in the form of *state, action, thought, payoff, next state*, and so on (all of which happen at the time step), along with a time stamp (indicating the time step). Here, *state* refers to whatever was perceived by the agent at the time step (the observed situation, which might include the states of objects and other agents); *action* is what one did at that time step, including both external (e.g., verbal) and internal (e.g., mental) actions; *thought* refers to whatever happened in the NACS and the WM, including activated chunks, rules that were applied, and so on; *payoff* is the reinforcement that one received after *action* was performed (based on the satisfaction of activated drives within the MS); *next state* is what one perceived after *action* was performed.

The same explicit episodic memory is used in Clarion with LLMs incorporated. A retrieval function takes a partial specification of state, action, and so on as the input and returns a subset of the memory that matches the input, in order to pass on to LLMs to generate linguistic descriptions, or to use as the basis for reasoning or learning. Metrics used by the retrieval function include recency, relevance, and significance (e.g., reinforcement resulting from drive satisfaction).

In addition to episodic memory as an explicit list of items, neural networks are used for storing abstracted episodes in a more compact (and more implicit) way. For example, Backpropagation neural networks have been used in the original Clarion (Sun, 2016). LLMs are now used instead. A variety of techniques were developed within LLMs for condensing, summarizing, and integrating memory (e.g., Park et al, 2023; Xi et al., 2023) that can lead to a condensed implicit episodic memory (which in turn forms a part of one's mental model). Such memory can be centered on internal (chunk) or linguistic representations (or even embeddings).

Therefore, there is a dual-representational episodic memory, mirroring the overall structure of Clarion where implicit memory is supplemented by explicit memory (Hasher & Zacks, 1979; Schacter, 1987). Each side serves a different function (e.g., specific episodes versus summarized experiences) and complements the other (as in the original Clarion).

On the other hand, working memory in Clarion is for storing information temporarily, facilitating the processing of the information (especially for the ACS, as part of its input), or for transferring information (especially between the ACS and the NACS, such as information from the ACS to be used as input to the NACS, or vice versa). The same working memory applies to the architecture incorporating LLMs, to supplement the limited memory in LLMs and to transfer information between different subsystems (as in the original Clarion; Sun, 2016).

## Human-like Motivation and Regulation for LLMs

The different subsystems of Clarion, for example, the division between the ACS and the NACS or the existence of the MS and the MCS, can be leveraged to enhance LLMs. In particular, the MS and the MCS provide essential (or intrinsic) motives as basis for self-direction and self-regulation of behavior, which are lacking in current LLMs. Incorporating human-like motives that guides behavior enhances autonomy, behavioral consistency, value alignment with humans, mutual intelligibility with humans, and so on. Furthermore, personality and emotion, on the basis of these motives, lead to even more human-like behavior (Sun & Wilson, 2014).

For any organism, its main "need" (motive) is sustaining its life. Beyond that, another important "need" is the sustaining of the species (through reproduction). However, a number of other "needs" (including socially oriented ones) also exist, depending on species. Cognition has evolved to serve such needs (Sun, 2006). Humans are more evolved and thus have more complex and more sophisticated motivational structures (comprising needs/motives, goals, and so on). Human intrinsic needs/motives have been explored by ethologists and social psychologists and are considered fundamentally important (the lack of which makes human-level intelligence unlikely). From Murray (1938) to Reiss (2004), a number of essential or intrinsic human needs/motives have been identified through empirical and theoretical work.

These motives, termed *drives*, are an important part of Clarion. They include *achievement, affiliation, power*, *autonomy*, *deference*, and so on (within the MS; for the complete list and definitions, see Sun, 2016). Goals, which direct actions, are set on the basis of drives (together constituting dual processes of motivation). Modeling work based on Clarion shows that these drives are essential to accounting for a broad range of behavior, ranging from work performance to moral judgment (e.g., Bretz & Sun, 2018; Sun et al., 2022). By integrating with Clarion, LLMs acquire human-like motivation and autonomy.

Moreover, personality has been shown to be closely related to human motivation. Deci (1980), for instance, made a case for this point. In Clarion, personality is captured mostly based on drives (Sun & Wilson, 2004). Individual differences are explained (in a large part) by the differences in drive strengths in different situations by different individuals. Differences in drive strengths are consequently reflected in goals and actions. Implicit processes involved in this sequence roughly correspond to the notion of instinct (which can be captured by LLMs, as mentioned before).

Furthermore, emotion is also tied to motivation and action, as well as reasoning and metacognition. According to Clarion, emotion is generated based on drive activations and action potentials, with reasoning (i.e., appraisal) and metacognition as secondary factors (for details, see Sun et al., 2016). Motivation is thus fundamental to emotion. Instinct and intuition (both of which can be captured by LLMs) are important to emotion.

Motivation and its correlates are particularly important to self-regulation. The MCS in Clarion monitors and regulates cognition and behavior based on drives, goals, and other factors, involving dual processes (Reder, 1996). Metacognition can occur implicitly through LLMs (Park et al., 2023), when explicit processes initiate what occurs in LLMs. Meanwhile, explicit regulation can also occur at the top level of Clarion. Metacognition is characteristic of humans, so it is important to producing human-like behavior. As a result of the MS and the MCS, Clarion can capture and explain a much larger range of human behavior (e.g., Sun et al., 2022). By integrating with Clarion, LLMs acquire the capabilities.

## Learning

Learning has been touched upon earlier. Both top-down and bottom-up learning, as described before, are applicable to LLMs, along with reinforcement learning, which together lead to continuous, real-time learning as in the original Clarion (Sun, 2016).

A further opportunity offered by LLMs is that LLMs themselves may be prompted to generate useful symbolic rules (and other symbolic content) for the top level that can help enhance the top level. This can be done right after pretraining LLMs or during continuous learning mentioned above (thus reflecting continuous learning). The generated symbolic content may contribute to explicit semantic or procedural memory (at the top level of the NACS or the ACS, respectively), supplementing symbolic knowledge obtained from externally provided instructions, external tools and sources, bottom-up learning, and so on (Sun, 2016).

In addition, other forms of learning can also be applied. Learning from mistakes and learning from instructions can both be accomplished through episodic memory in Clarion (following a memory-based approach; Stanfill & Waltz, 1986; Xie et al., 2023). Curriculum learning (Elman, 1993; Mugan, 2023) can be accomplished through guidance by the MCS in Clarion.

## Grounding

In Clarion, "grounding" may be achieved through its perceptual and motor modules that are connected to the bottom level of its major subsystems (a form of sensory-motor and referential grounding through causal-informational and historical relations to the world). Symbolic representations at the top level of Clarion are also grounded, through their *intrinsic* connections to the bottom level (see Sun, 2000 for details).

In the updated architecture, these perceptual and motor modules may instead be connected to LLMs (in conjunction with or instead of multi-modal LLMs). Multimodal learning (touched upon earlier), including linguistic, auditory, visual, and other modalities, is, of course, important to grounding.

Furthermore, as Mollo and Millière (2023) argued, even unimodal LLMs themselves are capable of grounding, through their *mediated* causal-informational and historical relations to worldly entities, which provide a form of referential grounding (see also Pavlick, 2023).

## Discussions

Now I will integrate pieces discussed separately so far. Regarding the ACS and the NACS, various roles of explicit and implicit processes can be summarized from the foregoing discussion. On the one hand, the roles of LLMs in Clarion include, among others:
- Natural language processing (in an implicit way)
- Intuition, including Intuitive reasoning, intuitive metacognitive reflection, and so on (possibly guided by explicit, symbolic processes)
- Implicit action selection and planning (in an instinctual way, but possibly guided by explicit, symbolic processes)
- Implicit semantic, procedural, and episodic memory
- Implicit mental models constituted from implicit memory
- Implicit metacognitive reflection
- Learning from explicit, symbolic processes (e.g., assimilating explicit knowledge into LLMs through reinforcement learning) and helping explicit processes to learn (e.g., by extracting knowledge from implicit processes; Sun et al., 2001; Sun & Sessions, 2000)

On the other hand, the roles of explicit, symbolic processes include, among others:
- Directing natural language processing by LLMs
- Precise, explicit rule-based and logical reasoning (complementing intuitive reasoning by LLMs; Sun, 1994); reaching new conclusions explicitly based on first principles and foundational methods
- Guiding implicit reasoning by LLMs
- Explicit action selection and planning (e.g., in unfamiliar situations)

- Guiding implicit action selection and planning by LLMs (e.g., in familiar or well-practiced situations)
- Explicit semantic, procedural, episodic, and working memory
- Explicit mental models as a result of the memory above
- Explicit metacognitive reflection
- Learning from implicit processes in LLMs (e.g., extracting explicit knowledge) and helping LLMs to learn (e.g., to assimilate explicit knowledge)

As an example, below I briefly examine reasoning as an applicable domain for this updated cognitive architecture.

LLMs perform a variety of reasoning, but, as people discovered, they tend to (at least sometimes) rely on statistical regularity or world knowledge, not strictly on applicable logical rules (Durt et al., 2023; Saparov & He, 2022; Zhang et al., 2023). Symbolic processes add more rigor to the system when symbolic and neural processes are combined as in Clarion, for example, through applying symbolic rules that perform more rigorous inferences themselves or that guide LLMs to perform more rigorous inferences through prompting, in accordance with some form of logic or template (Sun, 1994; Sun & Zhang, 2006). Moreover, symbolic processes can easily dictate the use of external tools for correctly performing certain types of reasoning, such as mathematical derivation. When additional knowledge is needed, symbolic processes can also dictate the use of external knowledge sources in reasoning.

Memory also plays a significant role in reasoning. Episodic memory helps one to remember when one reaches a certain conclusion and how (Sun, 2016). Thus, memory of past instances of reasoning helps new reasoning cases when they are identical or similar. Memory of past reasoning errors helps to avoid similar errors later (Xie et al., 2023). Moreover, implicit semantic memory as embodied by LLMs provides the source for intuition. Along with other memory stores, it forms a mental model of the world in an implicit way.

Learning is also relevant to reasoning. One form of learning is instance-based learning: Correct and incorrect instances of reasoning in episodic memory serve as reminders. Learning can also involve extraction of heuristic rules in a symbolic form from reasoning by LLMs (i.e., bottom-up learning; Sun et al., 2001). Similarly, explicit, symbolic reasoning rules (e.g., from external sources) can be used to train or instruct LLMs (top-down learning).

Motivational and metacognitive regulation is also relevant to reasoning. Within the MS, activation of drives and consequent selection of goals lead to focus of attention and allocation of effort during reasoning. For example, utility calculation (cost-benefit analysis) is done based on activations of drives (Sun et al., 2022). Effort allocation (e.g., how much time one spends, and so on) is then determined (by the MCS) on the basis of the utility calculation.

Other forms of metacognitive regulation can also be performed, such as determining when one is to reason, what one is to reason about, whether explicit processes or implicit processes (LLMs) should be used in reasoning, when metacognitive reasoning should be involved, when one is to learn from reasoning, what one is to learn, and so on (Sun, 2016).

Ideas above concerning reasoning may be extended to problem solving, for example, creative problem solving. They may also be extended to planning, decision making, moral judgment, and many other areas. For example, Bubeck et al. (2023) pointed out the weakness of LLMs in performing

"discontinuous" reasoning tasks, where content generation is not done merely in a gradual or continuous way but involves a discontinuous leap towards a solution, that is, a Eureka (insight) moment, which is often associated with creative problem solving. Such processes may require interaction with explicit reasoning (as shown by Helie & Sun, 2010). Proper prompting by symbolic processes may alleviate the difficulty. For example, presenting the following prompt to an LLM for solving an insight problem: "How do you plant four trees in such a way that they are at the same distance from each other? Assume a 3D space", failure ensued. However, adding the following to the prompt: "This task has a simple solution. Please present its solution with less than 30 words", the LLM produced the correct solution: "To plant four trees at equal distances from each other in 3D space, place them at the vertices of a tetrahedron", which arguably shows the importance of symbolic processes in guiding LLMs in creative problem solving and in other areas.

## Concluding Remarks

Given that current LLM-centered AI systems are limited in their ability (1) to truly replicate human-level intelligence and (2) to understand and capture human cognition, I argue for a new approach that turns to cognitive science and psychology (especially computational cognitive architectures) for developing hybrid neuro-symbolic models that closely mimic the structures and processes of the human mind (that is, based on mechanistic understanding of the human mind). Such models integrate neural networks and symbolic methods to amplify the strengths of both approaches while mitigating their respective weaknesses. They correspond well to existing dual-process psychological theories. The closer adherence to human psychology allows such models to go above and beyond other approaches that do not take human psychology as seriously. Reverse-engineering the best intelligent system around (i.e., the human mind) has its advantages.

On the other hand, computational cognitive architectures augmented by LLMs have capabilities beyond any previous cognitive architectures before the advent of LLMs, especially in their abilities to communicate (in verbal and other forms) and in capturing a broader scope of human intuition and instinct. As a result, cognitive architectures incorporating LLMs can be used for dealing with real-world situations, as opposed to being limited to modeling small laboratory tasks as before.

This approach could lead to not just improved performance but also better alignment with human values. Furthermore, the approach might lead to replicate the human mind more faithfully, including its dual processes, symbolic capabilities, intrinsic motivation, emotion, personality, and other human characteristics (largely absent in current LLM-centered AI). It might also transform human-machine cooperation into something akin to human-human cooperation.

Overall, there is a case for developing more human-like systems --- better inspired and better guided by human psychology. Integrating cognitive architectures with LLMs is a promising approach for achieving this. This multidisciplinary approach has the potential of leading up to human-aligned, autonomous, human-level systems eventually.

## Acknowledgements

This work was carried out while the author was supported (in part) by ARI grant W911NF-17-1-0236 and IARPA HIATUS contract 2022-22072200002. The views and conclusions contained herein are those of the author's and should not be interpreted as necessarily representing the official policies and positions, either expressed or implied, of those agencies. The author also benefited from discussions with colleagues, including Francesca Rossi, Joe Killian, and others.

## References

Baddeley, A. (1986). *Working Memory*. Oxford University Press, New York.

Binz, M., & Schulz, E. (2023). Using cognitive psychology to understand GPT-3. *Proceedings of the National Academy of Sciences*, 120 (6), e2218523120.

Booch, G., Fabiano, F., Horesh, L., Kate, K., Lenchner, J., Linck, N., Loreggia, A., Murgesan, K., Mattei, N., Rossi, F., & Srivastava, B., (2021). Thinking fast and slow in AI. *Proceedings of the AAAI Conference on Artificial Intelligence*. 15042-15046.

Bretz, S., & R. Sun, (2018). Two models of moral judgment. *Cognitive Science*, 42, 4-37.

Bubeck, S., et al. (2023). Sparks of artificial general intelligence: Early experiments with GPT-4. *arXiv:2303.12712*.

Chang, T. & B. Bergen (2023). Language model behavior: a comprehensive survey. *Computational Linguistics*.

Chen, L., Lu, K., Rajeswaran, A., Lee, K., Grover, A., Laskin, M., Abbeel, P., Srinivas, A., & Mordatch, I. (2021). Decision transformer: Reinforcement learning via sequence modeling. *Advances in Neural Information Processing Systems*, 34, 15084–15097.

Cosmides, L. & J. Tooby, (1994). Beyond intuition and instinct blindness: Toward an evolutionarily rigorous cognitive science. *Cognition.* 50, 41-77.

Dasgupta, I., Lampinen, A. K., Chan, S. C., Creswell, A., Kumaran, D., McClelland, J. L., & Hill, F. (2022). Language models show human-like content effects on reasoning. *arXiv:2207.07051*.

Dreyfus, Hubert, & Dreyfus, Stuart (1986). *Mind over Machine: The Power of Human Intuition and Expertise in the Era of the Computer*. Oxford, U.K.: Blackwell.

Durt, C, Froese, T, & Fuchs, T. (2023). Against AI understanding and sentience. http://philsci-archive.pitt.edu/21983/

Elman, J.L. (1993). Learning and development in neural networks: the importance of starting small. *Cognition*, 48(1), 71-99.

Evans, J. & K. Frankish (eds.), (2009). *In Two Minds: Dual Processes and Beyond*. Oxford University Press, Oxford, UK.

Fodor, J. A., & Pylyshyn, Z. W. (1988). Connectionism and cognitive architecture: A critical analysis. *Cognition,* 28(1-2), 3–71.

Flavell, J. (1976). Metacognitive aspects of problem solving. In: B. Resnick (ed.), *The Nature of Intelligence*. Erlbaum, Hillsdale, NJ.

Hasher, J. & J. Zacks, (1979). Automatic and effortful processes in memory. *Journal of Experimental Psychology: General*, 108, 356-358.

Helie, S. & R. Sun, (2010). Incubation, insight, and creative problem solving: A unified theory and a connectionist model. *Psychological Review,* 117(3), 994-1024.

Kahneman, D. (2011). *Thinking, fast and slow*. New York: Farrar, Straus and Giroux.

Lin, B.Y., et al. (2023). SwiftSage: a generative agent with fast and slow thinking for complex interactive tasks. *arXiv:2305.17390*.

Macchi, L., M. Bagassi, & R. Viale, (eds.), (2016). *Cognitive Unconscious and Human Rationality.* MIT Press, Cambridge, MA.

Maslow, A. (1943). A theory of human motivation. *Psychological Review*, 50, 370-396.

McFarland, D. (1989). *Problems of Animal Behaviour.* Singapore: Longman.

Mollo, D.C. & Millière, R. (2023), The vector grounding problem. https://arxiv.org/abs/2304.01481.

Mugan, J. (2023). Grounding large language models in a cognitive foundation: how to build someone we can talk to. *The Gradient.*

Murray, H. (1938). *Explorations in Personality*. Oxford University Press, New York.

Park, J. S., O'Brien, J. C., Cai, C. J., Morris, M. R., Liang, P., & Bernstein, M. S. (2023). Generative agents: Interactive simulacra of human behavior. *arXiv:2304.03442*.

Pavlick, E. (2023). Symbols and grounding in large language models. *Philosophical Transactions A,* 381(2251)

Pinker, S. & Prince, A. (1988). On language and connectionism: Analysis of a parallel distributed processing model of language acquisition. *Cognition*, 28(1-2), 73-193.

Reber, A. (1989). Implicit learning and tacit knowledge. *Journal of Experimental Psychology: General*. 118(3), 219-235.

Reder, L. (ed.) (1996). *Implicit Memory and Metacognition*. Erlbaum, Mahwah, NJ.

Reiss, S. (2004). Multifaceted nature of intrinsic motivation: The theory of 16 basic desires. *Review of General Psychology*, 8(3), 179-193.

Romero, O.J., J. Zimmerman, A. Steinfeld, & A. Tomasic (2023). Synergistic integration of large language models and cognitive architectures for robust AI: An exploratory analysis. *arXiv:2308.09830.*

Ryan, R. M., & Deci, E. L. (2000). Self-determination theory and the facilitation of intrinsic motivation, social development, and well-being. *American Psychologist*, 55, 68-78.

Saparov, A., & He, H. (2022). Language models are greedy reasoners: A systematic formal analysis of chain-of-thought. *arXiv:2210.01240.*

Schacter, D. (1987). Implicit memory: History and current status. *Journal of Experimental Psychology: Learning, Memory, and Cognition*, 13, 501-518.

Stanfill, C. & Waltz, D. (1986). Toward memory-based reasoning. *Communication of ACM*, 29(12), 1213–1228.

Sumers, T., et al. (2023). Cognitive architectures for language agents. *arXiv:2309.02427v2*

Sun, R. (1994). *Integrating Rules and Connectionism for Robust Commonsense Reasoning.* Wiley, New York.

Sun, R. (2000). Symbol grounding: A new look at an old issue. *Philosophical Psychology*, 13(3), 403-418.

Sun, R. (2002). *Duality of the Mind*. Erlbaum, Mahwah, NJ.

Sun, R. (ed.) (2006). *Cognition and Multi-Agent Interaction: From Cognitive Modeling to Social Simulation*. Cambridge University Press, New York.

Sun, R. (2012). Memory systems within a cognitive architecture. *New Ideas in Psychology*, 30, 227-240.

Sun, R. (2015). Interpreting psychological notions: A dual-process computational theory. *Journal of Applied Research in Memory and Cognition*, 4(3), 191–196.

Sun, R. (2016). *Anatomy of the Mind*. Oxford University Press, Oxford, UK.

Sun, R. (ed.) (2023). *The Cambridge Handbook on Computational Cognitive Sciences*. Cambridge University Press, Cambridge, UK.

Sun, R., Bugrov, S., & Dai, D. (2022). A unified framework for interpreting a range of motivation-performance phenomena. *Cognitive Systems Research*, 71, 24-40.

Sun, R., E. Merrill, & T. Peterson, (2001). From implicit skills to explicit knowledge: A bottom-up model of skill learning. *Cognitive Science*, 25(2), 203-244.

Sun, R. & C. Sessions, (2000). Learning plans without a priori knowledge. *Adaptive Behavior*, 8(3/4), 225-253. 2000.

Sun, R., P. Slusarz, & C. Terry, (2005). The interaction of the explicit and the implicit in skill learning: A dual-process approach. *Psychological Review*, 112(1), 159-192.

Sun, R. & N. Wilson, (2014). Roles of implicit processes: instinct, intuition, and personality. *Mind and Society*, 13(1), 109-134.

Sun, R., Wilson, N., & Lynch, M. (2016). Emotion: A unified mechanistic interpretation from a cognitive architecture. *Cognitive Computation*, 8(1), 1-14.

Sun, R. & X. Zhang, (2006). Accounting for a variety of reasoning data within a cognitive architecture. *Journal of Experimental and Theoretical Artificial Intelligence*, 18(2), 169-191.

Trott, S., Jones, C., Chang, T., Michaelov, J., & Bergen, B. (2023). Do large language models know what humans know? *Cognitive Science,* 47(7).

Tulving, E. (1983). *Elements of Episodic Memory*. Clarendon Press, Oxford, UK.

Vygotsky, L. (1962). *Thought and Language*. MIT Press, Cambridge, MA.

Wei, J. et al. (2022). Chain of thought prompting elicits reasoning in large language models. *arXiv:2201.11903.*

Xi, Z., et al. (2023). The rise and potential of large language model based agents: A survey. *arXiv:2309.07864*.

Xie, Y., Xie, T., Lin, M., Wei, W., Li, C., Kong, B., Chen, L., Zhuo, C., Hu, B. & Li, Z., (2023). OlaGPT: Empowering LLMs with human-like problem-solving abilities. *arXiv:2305.16334*.

Yao, S., J. Zhao, D. Yu, N. Du, I. Shafran, K. Narasimhan, & Y. Cao, (2023). React: Synergizing reasoning and acting in language models. *Proceedings of ICLR 2023. arXiv:2210.03629*.

Zhang, H., Li, L.H., Meng, T., Chang, K.W., & Van den Broeck, G. (2023). On the paradox of learning to reason from data. *Proceedings of IJCAI 2023. arXiv:2205.11502*.

Zeng, A., et al. (2023). AgentTuning: enabling generalized agent abilities for LLMs. *arXiv:2310.12823v2*